\RequirePackage{fix-cm}
\documentclass[twocolumn]{svjour3}          
\smartqed  

\usepackage{cite}
\usepackage[pdftex]{graphicx}
\usepackage{ragged2e}
\usepackage[tight,footnotesize]{subfigure}
\usepackage{rotating}
\usepackage{graphicx}
\usepackage{amsmath,amssymb} 
\usepackage{array}
\usepackage{multirow}
\usepackage{colortbl}
\usepackage{amsfonts}
\usepackage{pifont}
\usepackage{xspace}
\usepackage{etoolbox}
\usepackage{overpic}
\usepackage{color}
\usepackage{microtype}
\usepackage[title]{appendix}
\usepackage{xcolor}
\usepackage{booktabs}
\usepackage{colortbl}
\usepackage{bm}
\usepackage{makecell}

\newcommand{\orcid}[1]{\href{https://orcid.org/#1}{\includegraphics[width=10pt]{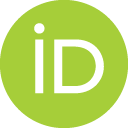}}}

\usepackage{hyperref}
\hypersetup{breaklinks=true,citecolor=blue, colorlinks}

\DeclareGraphicsExtensions{.pdf,.jpg,.png}

\usepackage{silence}
\journalname{Visual Intelligence}

\newcommand{\thename}[0]{ViTGaze}
\newcommand{\tablestyle}[2]{\setlength{\tabcolsep}{#1}\renewcommand{\arraystretch}{#2}\centering\footnotesize}

\begin{document}

\title{ViTGaze: Gaze Following with Interaction Features in Vision Transformers}

\titlerunning{ViTGaze}        

\author{Yuehao Song \orcid{0009-0003-2711-944X}        \and
  Xinggang Wang \orcid{0000-0001-6732-7823} \and 
  Jingfeng Yao \orcid{0009-0001-7775-0542} \and
  Wenyu Liu \orcid{0000-0002-4582-7488} \and
  Jinglin Zhang \orcid{0000-0002-6778-6664} \and
  Xiangmin Xu \orcid{0000-0003-4573-5820}
}


\institute{
Yuehao Song, Xinggang Wang, Jingfeng Yao, and Wenyu Liu are with School of Electronic Information and Communications, Huazhong University of Science and Technology, Wuhan, China. 
(Email: yhaosong@hust.edu.cn, xgwang@hust.edu.cn, jfyao@hust.edu.cn, liuwy@hust.edu.cn). \\
Jinglin Zhang is with School of Control Science and Engineering, Shandong University, Jinan, China. 
(Email: jinglin.zhang@sdu.edu.cn). \\
Xiangmin Xu is with School of Future Technology, South China University of Technology, Guangzhou, China. 
(Email: xmxu@scut.edu.cn). \\
Corresponding author: Xinggang Wang.
}

\date{Received: date / Accepted: date}

\maketitle

\begin{abstract}
Gaze following aims to interpret human-scene interactions by predicting the person's focal point of gaze.
Prevailing approaches often adopt a two-stage framework, whereby multi-modality information is extracted in the initial stage for gaze target prediction.
Consequently, the efficacy of these methods highly depends on the precision of the preceding modality extraction.
Others use a single-modality approach with complex decoders, increasing network computational load.
Inspired by the remarkable success of pre-trained plain vision transformers (ViTs),
we introduce a novel single-modality gaze following framework called \thename{}.
In contrast to previous methods, it creates a novel gaze following framework based mainly on powerful encoders (relative decoder parameters less than 1\%).
Our principal insight is that the inter-token interactions within self-attention can be transferred to interactions between humans and scenes.
Leveraging this presumption, we formulate a framework consisting of a 4D interaction encoder and a 2D spatial guidance module to extract human-scene interaction information from self-attention maps.
Furthermore, our investigation reveals that ViT with self-supervised pre-training has an enhanced ability to extract correlation information.
Many experiments have been conducted to demonstrate the performance of the proposed method.
Our method achieves state-of-the-art (SOTA) performance among all single-modality methods
(3.4\% improvement in the area under curve (AUC) score, 5.1\% improvement in the average precision (AP))
and very comparable performance against multi-modality methods with 59\% number of parameters less.


\end{abstract}

\section{Introduction}\label{sec1}
Gaze following is the task of predicting a person's gaze target in an image.
Specifically, given an image and a bounding box of a person's head, it aims to predict the location of the point where the person is watching.
It is widely applied in the fields of human-computer interaction~\cite{app_interact} and neuroscience~\cite{app_Neural}.
Gaze following through RGB images has been a longstanding topic of research, with numerous related studies developed over time.

Previous works have taken two approaches.
One approach introduces multi-modality frameworks to improve the prediction performance.
These methods often adopt a two-stage methodology.
In the initial stage, task-specific modality predictors are employed to extract supplementary information, including depth and poses, as additional inputs to compensate for the absence of human-scene interactions~\cite{Fang_DAM_2021_CVPR,Bao_ESC_2022_CVPR,Gupta_MM_2022_CVPR}.
In the second stage, the visual features and the multi-modality input extracted in the initial stage are combined and utilized by the decoder to regress the gaze target heatmap.
Another approach~\cite{Tu_HGGTR_2022_CVPR,Tonini_GOT_2023_ICCV,tu2023joint} involves a query-based decoder and utilizes additional object bounding boxes to improve performance by learning person-object interactions.
These methods employ a convolutional backbone for image feature extraction.
A transformer decoder~\cite{detr} is subsequently used to mix global information and provide gaze target predictions corresponding to the gaze queries.
However, both methods have drawbacks:
\renewcommand{\labelenumi}{\theenumi)}
\begin{enumerate}
    \item The additional information results in a multi-modal design with a two-stage framework. The accuracy depends on the performance of the prior predictions.
    \item The query-based methods require a heavy decoder, which increases the complexity of the whole design.
\end{enumerate}

We posit that these drawbacks stem from a shared design flaw: the absence of a sufficiently robust encoder for feature extraction of human-scene interactions.
Recently, the pre-trained plain vision transformers (ViTs)~\cite{vit} have demonstrated remarkable visual modeling capabilities.
A few works have explored the use of plain ViT on several downstream tasks~\cite{vitdet,vitpose,vitmatte} and achieved impressive results, highlighting the capacity of encoding rich features of the task-agnostic pre-trained ViT~\cite{dino,dinov2}.
Inspired by these previous works, it would be interesting to raise the following question: Can a pre-trained ViT provide an effective interactive representation between humans and the scene to describe a gazing relationship?

We propose \thename, a concise single-modality and lightweight gaze following method based on pre-trained plain ViTs.
Our principal observation is that the inter-token interactions within self-attention can be transferred to interactions between humans and scenes.
The dot product operation in the self-attention mechanism inherently encodes token correlations.
Therefore, we hypothesize that the interactions between humans and scenes are also embedded in these self-attentions. This assumption of the self-attention map is consistent with the observations of previous studies~\cite{LOST,weaktr}.
On this basis, we design a concise 4D interaction encoder to extract the interaction information from multi-level and multi-head attention maps and design a 2D spatial guidance module to guide it. 
Owing to the strong ability of pre-trained ViTs to extract interactions between objects, \thename{} does not introduce any decoder design.
In contrast to previous methods, \thename{} creates a brand-new paradigm of gaze following, which is mainly based on encoders (decoder parameters account for less than 1\%).
Furthermore, we also observe that self-supervised pre-trained ViTs are better at understanding token interactions. In contrast to previous ViT-based methods~\cite{vitdet,vitpose,vitmatte,cellvit}, our method further investigates the importance of attention maps while using the feature map for gaze following.

\begin{figure}[ht]
    \centering
    \includegraphics[width=\columnwidth]{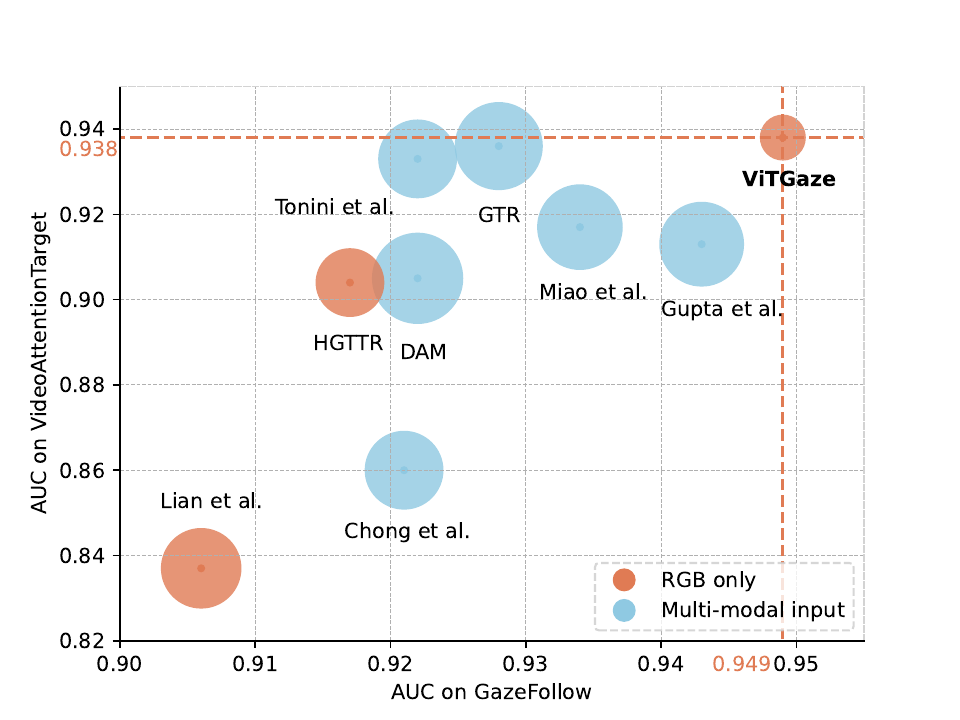}
    \caption{Comparison with state-of-the-art (SOTA) methods of area under curve (AUC) scores. \thename{} achieves SOTA performance among all methods in the AUC scores on both GazeFollow~\cite{Recasens_GazeFollow_2015_NIPS} and VideoAttentionTarget~\cite{Chong_VideoAttn_2020_CVPR}. The circle size indicates the number of parameters of each method. We compare our results (ViTGaze) with those of Lian et al.~\cite{Lian_ACCV_2019}, Chong et al.~\cite{Chong_VideoAttn_2020_CVPR}, DAM~\cite{Fang_DAM_2021_CVPR}, HGTTR~\cite{Tu_HGGTR_2022_CVPR}, Tonini et al.~\cite{Tonini_GOT_2023_ICCV}, GTR~\cite{tu2023joint}, Miao et al.~\cite{Miao_PDP_2023_WACV}, and Gupta et al.~\cite{Gupta_MM_2022_CVPR}.}
    \label{fig:bubbles}
\end{figure}

Compared with existing state-of-the-art methods, our method has advantages in both performance and efficiency.
We conduct experiments on the two most widely-used benchmarks, the GazeFollow~\cite{Recasens_GazeFollow_2015_NIPS} and VideoAttentionTarget~\cite{Chong_VideoAttn_2020_CVPR}.
The experimental results demonstrate the superiority of our method.
Specifically, our method achieves a 3.4\% improvement in the area under curve (AUC) and a 5.1\% improvement in the average precision (AP) among single-modality methods, resulting in a new state-of-the-art (SOTA) performance as shown in Fig.~\ref{fig:bubbles}.
In addition, our method achieves comparable performance to multi-modal methods (1.8\% lower in distance error but 2.7\% higher in AUC) with 59\% fewer parameters. These results provide ample evidence of the validity of our method, demonstrating that a single-modality lightweight framework built upon pre-trained ViTs could also achieve SOTA performance in gaze following.

Our contributions can be summarized as follows:
\renewcommand{\labelenumi}{\theenumi)}
\begin{enumerate}
    \item We propose \thename{}, a single-modality lightweight gaze following framework based on pre-trained vision transformers.
    To the best of our knowledge, this is the first gaze-following method built upon the pre-training of ViTs.
    \item We demonstrate the feasibility of extracting human-scene interaction information from inter-tokens interactions in self-attentions and design a 4D interaction module guided by a 2D spatial guidance module.
    \item We evaluate our method on GazeFollow~\cite{Recasens_GazeFollow_2015_NIPS} and VideoAttentionTarget~\cite{Chong_VideoAttn_2020_CVPR} benchmarks. 
    Our method achieves SOTA performance (3.4\% improvement in the AUC and 5.1\% improvement in the AP) among the single-modality methods and very comparable performance with multi-modality methods with 59\% fewer parameters.
\end{enumerate}

\section{Related Work}\label{sec2}
\subsection{Gaze Following}
Research on gaze behavior has drawn significant academic interest across multiple domains.
One prominent area is gaze estimation~\cite{zhong2024uncertainty}, which involves inferring gaze direction from facial cues.
Another key domain is scan path prediction~\cite{zhong2024spformer,xia2020evaluation}, which focuses on modeling the sequence of fixations across an image.
Unlike these works, gaze following~\cite{Recasens_GazeFollow_2015_NIPS} aims to interpret human-scene interactions by locating the gaze target of a person to provide an understanding of how individuals interact with their surroundings.
Previous research on gaze following can be classified into two principal categories: single-modality methods and multi-modality methods.

\paragraph{Methods with single modality.}
The ground-breaking work~\cite{Recasens_GazeFollow_2015_NIPS} designs a single modality architecture using only RGB images as the input.
In this work, gaze following is formulated as the combination of the individual's gaze field and the global saliency map~\cite{saliency1,saliency2,saliency3}.
The subsequent series of works~\cite{Chong_Connect_2018_ECCV,Lian_ACCV_2019,zhaoLearningDrawSight2020} focus on providing human-scene interactions through auxiliary tasks such as sight line estimation~\cite{Lian_ACCV_2019,zhaoLearningDrawSight2020}.
GaTector~\cite{Wang_GaTector_CVPR_2022} proposes a unified framework to extract features from both head crops and scenes with a shared backbone.
HGTTR~\cite{Tu_HGGTR_2022_CVPR} proposes a transformer-based encoder-decoder architecture, which provides implicit interaction clues through the global modeling of self-attention.
Despite the concise input of these methods, they often achieve unsatisfactory performance compared with multi-modality methods.

\paragraph{Methods with multiple modalities.}
To further improve prediction performance, a few methods utilize additional modality information besides RGB images.
Fang et al.~\cite{Fang_DAM_2021_CVPR} proposed a depth-assisted architecture to estimate spatial relationships between the person and the scene through an extra depth estimator. 
Furthermore, the human pose estimation sub-task is incorporated into the framework~\cite{Bao_ESC_2022_CVPR,Gupta_MM_2022_CVPR,Jian_Enhanced_ICMM_2020}, offering a more precise 3D spatial interaction estimation.
The temporal information is utilized in Refs. \cite{Chong_VideoAttn_2020_CVPR,Miao_PDP_2023_WACV} to improve prediction performance.
Additional object segmentation masks are incorporated in Refs. \cite{Chen_TPNet_TCSVT_2022,Hu_GazeTargetEstimation_TCSVT_2022} to model object interactions.
Samy et al.~\cite{sharingan} directly embeded the head position along with the coarse sight direction into a token and utilized a ViT to predict precise gaze targets. 
GTR~\cite{tu2023joint} incorporates gaze object detection into the HGTTR framework, providing object-level supervision for accurate gaze localization.
Francesco et al.~\cite{Tonini_GOT_2023_ICCV} expanded DETR~\cite{detr} to detect objects and gaze points simultaneously.
However, these methods mostly adopt a two-stage design and their performance is dependent on the prediction results from the first stage.

Hence, how to design a concise and high-performance single-modality gaze following framework remains an unsolved problem.

\subsection{Pre-training of Vision Transformers}
\paragraph{Self-supervised learning.}
Pre-training can improve ViT~\cite{vit} in learning transferable visual representations.
Recently, researchers~\cite{beit,mae} focus on self-supervised pre-training employing masked image modeling (MIM).
Furthermore, the integration of CLIP~\cite{clip} with MIM by Refs.~\cite{mvp,cae,eva,eva02} results in a notable enhancement in the performance of pre-training.
Notably, approaches such as those in Refs.~\cite{zhou2021ibot,dinov2} leverage online tokenizers to further optimize the performance of pre-training.

\paragraph{Downstream tasks.}
The pre-trained representations of ViTs can significantly enhance the performance of downstream tasks.
ViTDet~\cite{vitdet} applies pre-trained ViT to object detection by constructing a simple feature pyramid.
ViTPose~\cite{vitpose} investigates the performance of ViT in pose estimation tasks via a similar approach.
Additionally, ViT is employed in ViTMatte~\cite{vitmatte} in image matting tasks by designing a detail capture module.
In the medical field, pre-trained ViT can also further enhance the performance of cell segmentation~\cite{cellvit}.
In contrast, WeakTr~\cite{weaktr} leverages attention maps from a pre-trained ViT for weakly supervised image segmentation.

However, it remains an unexplored question whether the pre-trained representations contain enough interaction information to facilitate gaze following.

\section{Method}\label{sec3}

\subsection{Preliminary: Self-Attention}
We first review the dot-product self-attention mechanism proposed in Ref.~\cite{transformer}.
The input token sequence $\bm{x}\in \mathbb{R}^{L\times C}$ is first transformed into keys, queries, and values, where $L$ is the length of the sequence and $C$ is the number of channels.
Then the dot product performed between queries and keys captures token correlations which are normalized and used as weights to aggregate the values.
\begin{equation}
    \bm{A} = \sigma(\frac{\bm{q}\bm{k}^\mathrm{T}}{\sqrt{C}}),
\end{equation}
\begin{equation}
    \bm{v}^\mathrm{out} = \bm{A}\bm{v},
\end{equation}
where $\bm{q}$, $\bm{k}$, and $\bm{v}$ are queries, keys, and values respectively, and $\sigma$ refers to the softmax operation.
The normalized dot product, which is also known as the attention map $\bm{A}\in \mathbb{R}^{L\times L}$, adaptively determines the relationship between each pair of tokens.
In the vision region, an image with a size of $h\times w$ is first flattened to a patch sequence and then generates an attention map $\bm{A}_\mathrm{p}\in\mathbb{R}^{hw\times hw}$, which inherently follows a 4D paradigm that effectively indicates the interactions among patches.
This 4D interaction feature map has proven to be particularly effective in capturing patch correlations in an image, resulting in advanced performance in various computer vision tasks, such as weakly-supervised segmentation~\cite{weaktr} and unsupervised object localization~\cite{LOST}.

\subsection{Overall Structure}
\begin{figure*}[t]
    \centering
    \includegraphics[width=\textwidth]{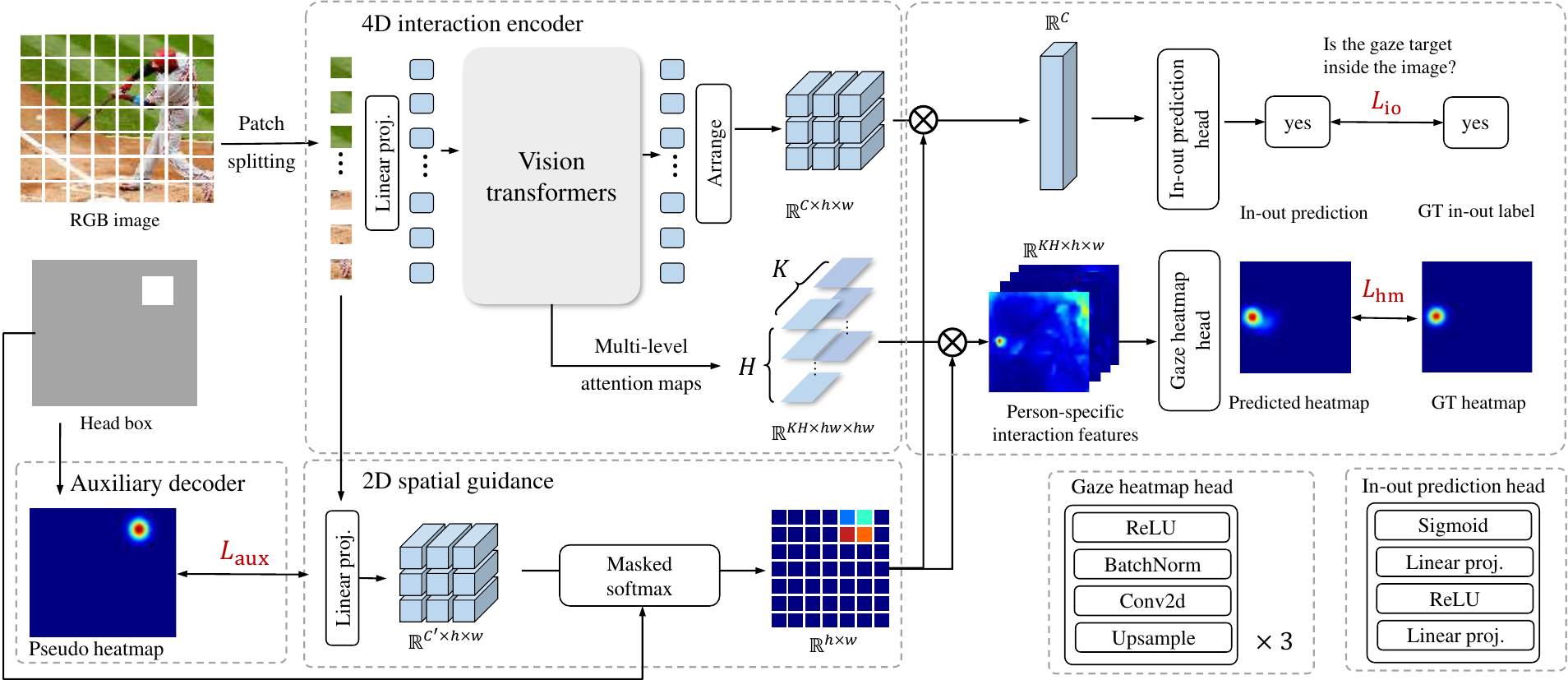}
    \caption{Overall structure of \thename{}. We achieve high-performance gaze following by predicting interactions with multi-level and multi-head attention maps, which we refer to as 4D features, guided by 2D spatial information. 
    It leverages the pre-trained vision transformer and lightweight decoders which have fewer than 1 M parameters.
    $\bigotimes$ refers to the weighted sum. $C, K, H, h, w$ refers to the number of feature channels, the number of semantic levels, the number of attention heads, and the height and the width of the input image, respectively.
    $L_\mathrm{hm}$, $L_\mathrm{io}$, and $L_\mathrm{aux}$ refer to gaze heatmap loss, gaze in-out loss, and auxiliary head regression loss.
    Linear proj., Conv2d, BatchNorm, ReLU, Sigmoid, and Upsample refer to linear projection, 2D convolution, batch normalization, ReLU activation, sigmoid activation, and bilinear upsample, respectively.}
    \label{fig:overall}
\end{figure*}

The overall structure of our method is shown in Fig.~\ref{fig:overall}.
It takes an image and a bounding box of the person's head as the inputs.
It outputs a heatmap that locates the gaze target with the highest response, and the probability that the person watches outside the scene.

Our method is composed of three components: a pre-trained vision transformer encoder that extracts a feature map with rich semantics and multi-level 4D interaction features, a 2D spatial guidance module to determine the spatial weights corresponding to each patch for person-specific interaction feature aggregation, and two prediction heads for heatmaps and in-out prediction.
These components are described in detail in the following subsections.

\subsection{Multi-level 4D Interaction Encoder}
\begin{figure*}[t]
    \centering
    \includegraphics[width=0.85\textwidth]{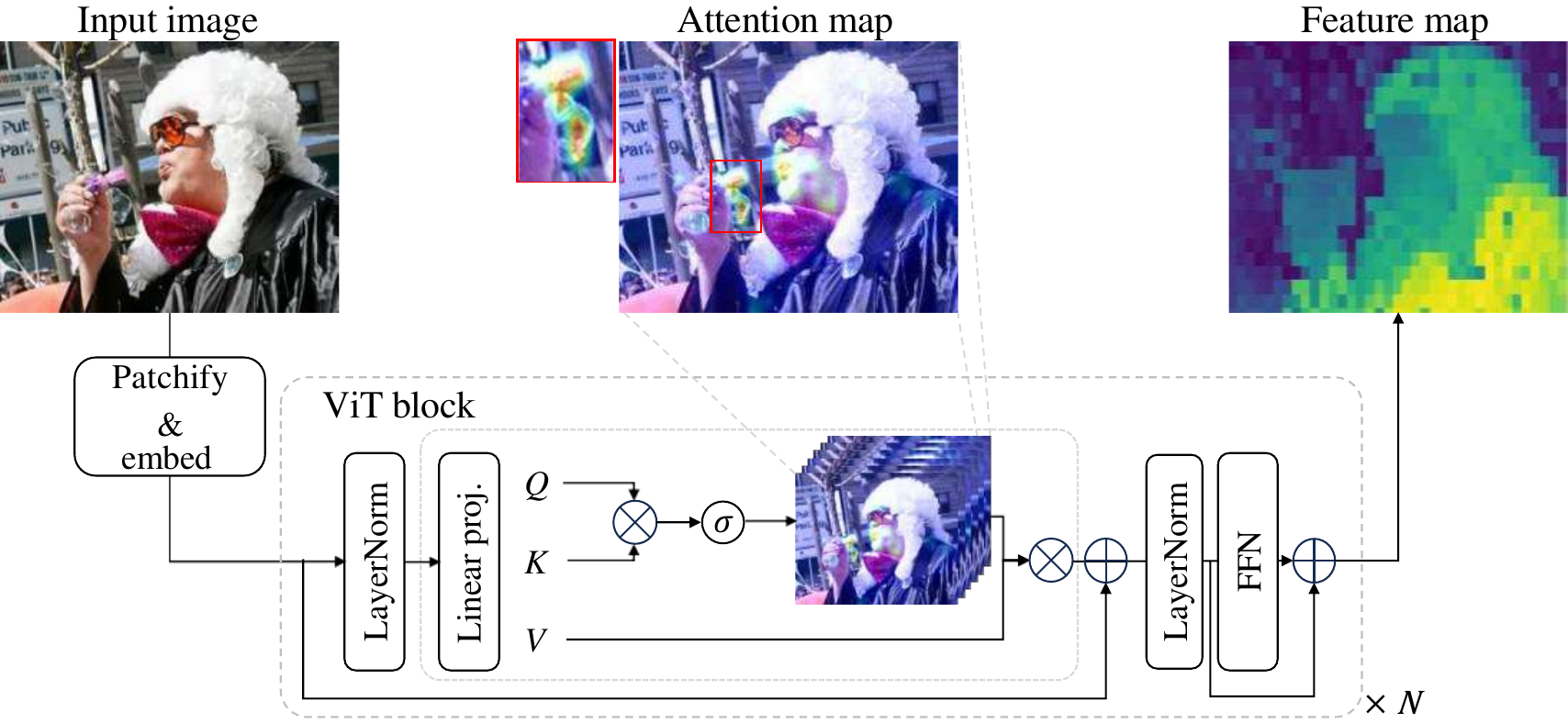}
    \caption{Visualization of vision transformer (ViT) features.
    In contrast to the feature map revealing global object-level semantics, the attention map of tokens overlapped with the head reflects human-scene interactions.
    Q, K, V, and N refer to queries, keys, values, and the number of transformer blocks.
    $\bigotimes$, $\bigoplus$, and $\sigma$ refer to weighted sum, add, and softmax.
    LayerNorm and FFN refer to layer normalization and the feed-forward network.}
    \label{fig:vit_features}
\end{figure*}

\begin{figure*}[t]
    \centering
    \subfigure[Our design]{
        \includegraphics[width=0.43\textwidth]{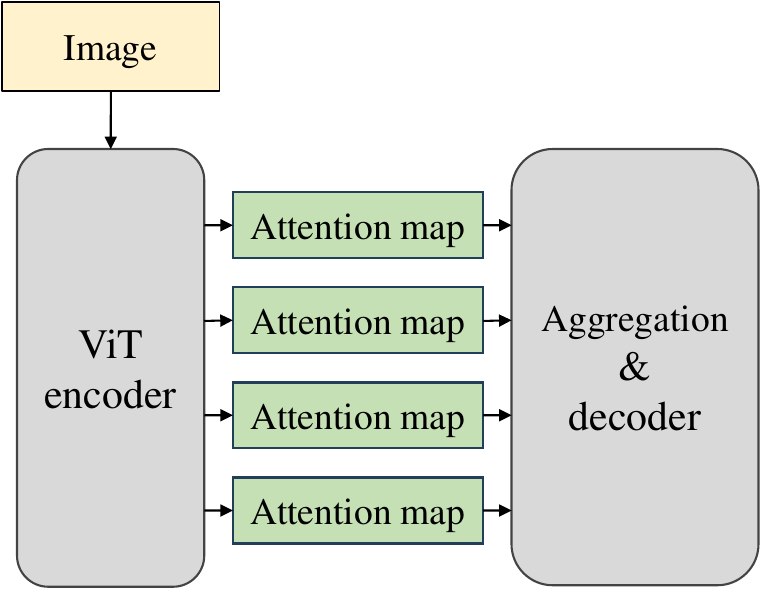}
        \label{fig:my_encoder}
    }
    \subfigure[Previous tasks]{
        \includegraphics[width=0.43\textwidth]{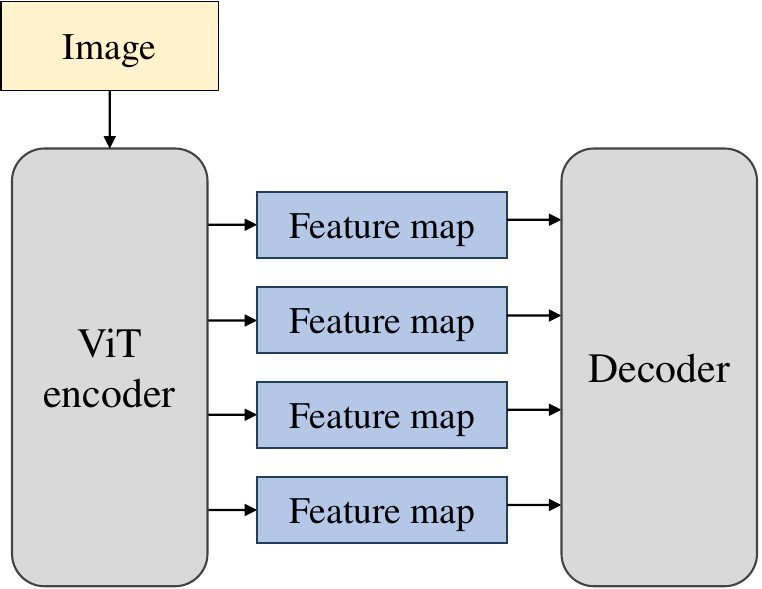}
        \label{fig:prev_encoder}
    }
    \caption{Comparisons with previous ViT-based tasks. In contrast to previous tasks that use feature maps, \thename{} leverages interaction information encoded in the attention maps.}
    \label{fig:encoders}
\end{figure*}

Inspired by the capacity of self-attention to capture patch-level interactions, we propose a 4D interaction encoder for efficient human-scene interaction estimation.
In Fig.~\ref{fig:vit_features}, we visualize the attention map of the token overlapped with the head of a person and the feature map output by the last block of a DINOv2~\cite{dinov2} pre-trained ViT.
Rich semantic information in the feature map enables effective distinction between objects.
However, it cannot represent interactions between image regions.
In contrast, a specific token's attention map is capable of acquiring interaction information between it and other regions.
Therefore, we propose the interaction encoder to leverage attention maps, which are referred to as 4D interaction features, from pre-trained ViTs with a simple adaptation (Fig.~\ref{fig:my_encoder}).
Compared with the encoders used in the previous tasks (Fig.~\ref{fig:prev_encoder}) such as object detection~\cite{vitdet} and image matting~\cite{vitmatte} that utilize feature maps to distinguish objects, the interaction feature-based encoder inherently facilitates the explicit capture of patch relations, which is required by gaze following.

Given an image with a resolution of $H_\mathrm{in}\times W_\mathrm{in}$. The interaction encoder extracts 4D features, which explicitly describe the interactions between image patches. The correlations among them are represented as a 4D tensor with a size of $h\times w\times h\times w$, where $h=H_\mathrm{in}/P$ and $w=W_\mathrm{in}/P$, and $P$ denotes the patch-size of the ViT.
Different from the final feature map leveraged in other regions~\cite{vitdet,vitpose,vitmatte}, this 4D representation reflects the inner-token relations which are more effective in gaze following.

Transformers pre-trained with masked image modeling capture correlations at multiple scales in multiple layers~\cite{dark_secret}.
To this end, we extract multi-level and multi-head 4D features to capture correlations between tokens with multiple distances.
Specifically, we extract features from $K=4$ transformer layers, and each feature is divided into $H=6$ sub-features based on different attention heads.
These attention maps represent patch interactions at both local and global levels and are combined to create multi-level 4D interaction features $\bm{A}_\mathrm{pm}\in \mathbb{R}^{KH\times h\times w\times h\times w}$, where $K$ and $H$ refer to the number of semantic levels and the number of heads in the multi-head self-attention, respectively.
The extracted features are then guided by spatial information and used for heatmap prediction.

\subsection{2D Spatial Guidance}
We propose a 2D spatial guidance module for aggregating the 4D interaction features with the head position to obtain person-specific interaction features.
The insight between this module has two parts:
1) The attention map describes how each image patch interacts with all other patches in a 4D paradigm, while the gaze feature for following the specific person's gaze must be a 2D feature map to predict the spatial distribution of gazing probability.
2) Due to the rich interaction features being fully extracted by the ViT encoder, it is unnecessary to design a heavy branch to extract head features and gaze cones, which is widely adopted in the previous literature~\cite{Recasens_GazeFollow_2015_NIPS,Miao_PDP_2023_WACV,Tonini_GOT_2023_ICCV,Gupta_MM_2022_CVPR}.
Therefore, we formulate the aggregation as a simple weighted sum of 4D interaction features in two spatial dimensions on the basis of the head position and head features.
The module is constructed with a simple two-layer multi-layer perceptron (MLP) with softmax activation, which is used to calculate the weights of each patch.
Background patches are masked before the softmax operation to guarantee that the weights are unique to the target individual.

Built upon this 2D spatial guidance, we proceed to feature aggregation with guidance. 
We obtain the person-specific interaction features $\bm{F}_\mathrm{pi}$ through a weighted sum of 4D interaction features $\bm{A}_\mathrm{pm}\in\mathbb{R}^{KH\times h\times w\times h\times w}$ with 2D guidance $\bm{G}\in\mathbb{R}^{h\times w}$ as weights. 
This can be expressed in the form of matrix multiplication as Eq.~(\ref{eq:interaction-feature}).

\begin{equation}
    \bm{F}_\mathrm{pi}=\bm{A}_\mathrm{pm}\bm{G}.
    \label{eq:interaction-feature}
\end{equation}

The output interaction features $\bm{F}_\mathrm{pi}\in\mathbb{R}^{KH\times h\times w}$ represent multi-level person-specific interactions between the target person and each patch.
With this simple aggregation, we transfer the abundant interaction information in the whole image to the person-scene interaction corresponding to the specific person.

Unlike the interaction feature, the in-out prediction feature focuses more on the global semantics instead of the geometric comprehension of the image,
which is provided in the final ViT feature maps.
Therefore, we use image token features $\bm{F}_\mathrm{io}\in\mathbb{R}^{KH}$, guided by 2D guidance.
It is expressed by Eq.~(\ref{eq:in-out-feature}),
while $\bm{F}_\mathrm{vit}\in\mathbb{R}^{KH\times h\times w}$ represents the image token features from the last layer of ViT.

\begin{equation}
    \bm{F}_\mathrm{io}=\bm{F}_\mathrm{vit}\bm{G}.
    \label{eq:in-out-feature}
\end{equation}

To further engage the module in feature extraction of heads, we introduce an auxiliary head that predicts whether a patch overlaps with the heads of any people.
The auxiliary head shares the stem with the main branch and introduces an extra prediction layer to predict the patch-level probability of overlapping with the head of a person to provide more supervision of head features.
Notably, in gaze-following settings, this auxiliary task does not require any supplementary input modalities such as depth, human poses, or object segmentation.

\subsection{Prediction Heads}
This component uses the person-specific interaction feature and the in-out prediction feature to predict a gaze target heatmap $\bm{H}^\mathrm{pr}_\mathrm{g}$ and a value $P_\mathrm{o}$ indicating the probability that the person watches outside the scene.
These two parts are detailed below.

\textbf{Gaze heatmap head.}
This module employs three groups of bilinear interpolation and convolutional layers to predict the gaze location. 
It converts person-specific interaction features to a logit map, where the highest activation point refers to the predicted gaze target.

\textbf{In-out prediction head.}
We design an MLP head to predict whether the person's gaze point is located in the image.
The head vector is fed into the two-layer MLP followed by sigmoid activation to obtain the probability that the person watches outside.

\subsection{Training objective}\label{loss}
We train our model in an end-to-end paradigm with the training objective as a weighted sum of all tasks.
The loss consists of three parts: gaze heatmap loss, gaze in-out loss, and auxiliary head regression loss.

Gaze heatmap loss $\mathcal{L}_\mathrm{hm}$ measures the error of gaze target prediction with a mean square error between the predicted gaze heatmap and the ground truth generated by a Gaussian blob centered at the coordinate of the gaze target.

Gaze in-out loss $\mathcal{L}_\mathrm{io}$ measures the prediction error of whether the person watches outside.
It is achieved by a focal loss~\cite{focal_loss} between the predicted probability and the ground truth label.

Auxiliary head regression loss $\mathcal{L}_\mathrm{aux}$ is designed to constrain the 2D spatial guidance.
It is defined as the binary cross-entropy loss between the predicted head occurrence and the ground truth heatmap which is a combination of Gaussian blobs centered at head bounding boxes in annotations.

The final loss is a linear combination of the three losses:

\begin{equation}
\mathcal{L}=\lambda_{1}\mathcal{L}_\mathrm{hm}+\lambda_{2}\mathcal{L}_\mathrm{io}+\lambda_{3}\mathcal{L}_\mathrm{aux},
\end{equation}
where $\lambda_1$, $\lambda_2$, and $\lambda_3$ are weights of $\mathcal{L}_\mathrm{hm}$, $\mathcal{L}_\mathrm{io}$, and $\mathcal{L}_\mathrm{aux}$.

\section{Experiment}\label{sec4}
\subsection{Datasets and Evaluation Metrics}
\textbf{Datasets.}
We train and test \thename{} on the Gazefollow~\cite{Recasens_GazeFollow_2015_NIPS} dataset and VideoAttentionTarget~\cite{Chong_VideoAttn_2020_CVPR} dataset.

Gazefollow~\cite{Recasens_GazeFollow_2015_NIPS} contains over 130 K annotations of people and their gaze targets in 122 K images.
The dataset is annotated with head bounding boxes, gaze points, and in-out-of-frame labels provided by Ref.~\cite{Chong_Connect_2018_ECCV}.
GazeFollow focuses on gaze targets inside the images, therefore only gaze heatmap regression is evaluated.

VideoAttentionTarget~\cite{Chong_VideoAttn_2020_CVPR} contains 1331 video clips with 165 K annotations.
We evaluate both gaze heatmap regression and watching-outside prediction.

\textbf{Evaluation metrics.}
We adopt the following metrics to evaluate the performance of the proposed method.
AUC reflects the prediction confidence of gaze heatmaps.
Distance (Dist.) refers to the relative Euclidean distance between the ground truth and the predicted position with the highest confidence.
Since 10 annotations are provided in GazeFollow for each instance, we report both the minimum distance (Min. Dist.) and average distance (Avg. Dist.).
We use AP to evaluate the performance for watching-outside prediction in VideoAttentionTarget.

\subsection{Implementation Details}
~~~~\textbf{Model structure.}
We adopt ViT-S~\cite{vit} as the transformer encoder.
The encoder consists of 12 blocks with a multi-head self-attention of 6 heads.
The multi-level 4D interaction is constructed with attention maps from the 3-rd, 6-th, 9-th, and 12-th blocks.
We use an efficient implementation of multi-head attention which is available in the xFormers~\cite{xFormers2022} library.

\textbf{Unbiased data processing.}
The coordinate encoding process (i.e. transforming ground-truth coordinates to heatmaps) used in current methods introduces quantization error of gaze targets and thus degrades model performance.
To address this dilemma, we follow DARK\cite{darkpose} to generate gaze heatmaps with a Gaussian kernel using the real gaze target as the center without quantization of the center coordinates.
We also adopt the post-processing method proposed in Ref.~\cite{darkpose} to infer the final gaze target location.

\textbf{Model training.}
For GazeFollow~\cite{Recasens_GazeFollow_2015_NIPS}, we initialize the model with weights of DINOv2~\cite{dinov2}.
The model is trained for 15 epochs via the AdamW~\cite{adamw} optimizer with a batch size of 48.
The initial learning rate is set as $0.01$, and decays to $0.001$ with a cosine scheduler.
The weight decay is set as 0.1.
We follow DINOv2~\cite{dinov2} to increase the resolution of images to $518\times 518$ in the last epoch.
In addition to basic data augmentations, we follow DINOv2~\cite{dinov2} to apply random masks to the images.
Specifically, we replace patches in the background that occupy less than 50\% of the area with a mask token with a probability of 0.5.
The loss coefficients are $\gamma$ = 2 in the focal loss, $\lambda_{1}$ = 100 for the heatmap regression, $\lambda_{2}$ = 1 for the watching-outside prediction, and $\lambda_{3}$ = 1 for auxiliary head regression.
For VideoAttentionTarget, following Refs.~\cite{Fang_DAM_2021_CVPR,Gupta_MM_2022_CVPR}, we finetune the model initialized with weights learned from GazeFollow.
We train for 1 epoch with the learning rate of $1.0\times 10^{-4}$ on VideoAttentionTarget.
The detailed training configurations are outlined in Appendix~\ref{sec:settings}.

\subsection{Comparison with State-of-the-Art}
We present the quantitative results for the GazeFollow and VideoAttentionTarget datasets in Tab.~\ref{tab:sota}.
To ensure a fair comparison, we also annotate the modalities used in each model.
Additionally, we include the parameter counts and floating point operations (FLOPs) of each model to facilitate the comparison of their efficiency, where the additional modality extractors required by multi-modality methods are not taken into account.

\begin{table*}[t]
    \centering
    \tablestyle{7.5pt}{1.0}
    \caption{Benchmark on two datasets.
    \thename{} achieves state-of-the-art (SOTA) performance among single-modality methods.
    Additionally, it performs on par with preceding multi-modality approaches with 59\% fewer parameters and 25\% lower computational cost.
    The best and second-best results are highlighted in bold and underlined.
    \#Params.: number of parameters; FLOPs: floating point operations; AUC: area under curve; Avg. Dist.: average distance; Min. Dist.: minimum distance; AP: average precision.
    Upward arrows indicate better performance with higher values, while downward arrows indicate the opposite.}
    \begin{tabular}{lccrrrrrr}
        \toprule
        \multirow{2.5}{*}{Method} & \multirow{2.5}{*}{\#Params.} & \multirow{2.5}{*}{\makecell[c]{GFLOPs\\(224$\times$224)}} & \multicolumn{3}{c}{GazeFollow~\cite{Recasens_GazeFollow_2015_NIPS}} & \multicolumn{3}{c}{VideoAttentionTarget~\cite{Chong_VideoAttn_2020_CVPR}} \\
        \cmidrule(lr){4-6}  \cmidrule(lr){7-9}
         & & & AUC$\uparrow$ & Avg. Dist.$\downarrow$ & Min. Dist.$\downarrow$ & AUC$\uparrow$ & Dist.$\downarrow$ & AP$\uparrow$ \\ 
        \midrule
        \multicolumn{9}{c}{multi-modality methods} \\
        \midrule
        Chong et al.~\cite{Chong_VideoAttn_2020_CVPR} & 54 M & 9.08 & 0.921 & 0.137 & 0.077 & 0.860 & 0.134 & 0.853 \\
        DAM~\cite{Fang_DAM_2021_CVPR} & 69 M & 6.53 & 0.922 & 0.124 & 0.067 & 0.905 & 0.108 & 0.896 \\
        TPNet~\cite{Chen_TPNet_TCSVT_2022} & - & - & 0.908 & 0.136 & 0.074 & - & - & - \\
        VSG-IA~\cite{Hu_GazeTargetEstimation_TCSVT_2022} & 57 M & 9.74 & 0.923 & 0.128 & 0.069 & 0.880 & 0.118 & 0.881 \\
        ESCNet~\cite{Bao_ESC_2022_CVPR} & - & - & 0.928 & 0.122 & - & 0.885 & 0.120 & 0.869 \\
        Gupta~\cite{Gupta_MM_2022_CVPR} & 61 M & 5.32 & \underline{0.943} & 0.114 & 0.056 & 0.913 & 0.110 & 0.879 \\
        GTR~\cite{tu2023joint} & - & - & 0.928 & 0.114 & 0.057 &\underline{0.936} & \textbf{0.095} & \underline{0.927} \\
        Miao et al.~\cite{Miao_PDP_2023_WACV} & 62 M & 9.13 & 0.934 & 0.123 & 0.065 & 0.917 & 0.109 & 0.908 \\
        Tonini et al.~\cite{Tonini_GOT_2023_ICCV} & 54 M & 5.78 & 0.922 & \textbf{0.069} & \textbf{0.029} & 0.933 & 0.104 & \textbf{0.934} \\
        \midrule
        \multicolumn{9}{c}{single-modality methods} \\
        \midrule
        Recasens et al.~\cite{Recasens_GazeFollow_2015_NIPS} & - & - & 0.878 & 0.190 & 0.113 & - & - & - \\
        Chong et al.~\cite{Chong_Connect_2018_ECCV} & - & - & 0.896 & 0.187 & 0.112 & 0.830 & 0.193 & 0.705 \\
        Lian et al.~\cite{Lian_ACCV_2019} & 56 M & 10.65 & 0.906 & 0.145 & 0.081 & 0.837 & 0.165 & - \\
        HGTTR~\cite{Tu_HGGTR_2022_CVPR} & 43 M & 4.89 & 0.917 & 0.133 & 0.069 & 0.904 & 0.126 & 0.854 \\
        \rowcolor{gray!20}
        \thename{} (Ours) & \textbf{22 M} & 4.62 & \textbf{0.949} & \underline{0.105} & \underline{0.047} & \textbf{0.938} & \underline{0.102} & 0.905 \\
        \bottomrule
    \end{tabular}
    \label{tab:sota}
\end{table*}

\textbf{Accuracy.}
As demonstrated in Tab.~\ref{tab:sota}, we compare \thename{} with previous multi-modality and single-modality methods.
\thename{} achieves new SOTA performance among single-modality methods.
Compared with the previous SOTA method~\cite{Tu_HGGTR_2022_CVPR}, \thename{} outperforms 3.4\% in terms of the AUC in GazeFollow and 5.1\% in terms of AP in VideoAttentionTarget.
This improvement demonstrates that \thename{} is efficient in leveraging single-modality data to achieve better performance.
Additionally, our method shows performance that is comparable to multi-modality methods without extra input used during prediction.
For instance, compared with the previous SOTA multi-modality method~\cite{Tonini_GOT_2023_ICCV}, our method is only 1.8\% lower in terms of the distance but achieves 2.7\% higher AUC on the GazeFollow benchmark.
This indicates that even with fewer modalities, \thename{} can match and even exceed the performance of complex multi-modality methods.
This highlights the efficiency and robustness of \thename{} to extract and utilize interaction features from single-modality data.

\textbf{Efficiency.}
As displayed in Tab.~\ref{tab:sota}, our method achieves the SOTA performance with only 22 M parameters and a computational cost of only 4.62 GFLOPs.
Even the computational overhead associated with the modality extraction stage required by multi-modality methods is not considered, our approach still achieves a reduction of over 50\% in the number of parameters and over 25\% in the computational demand in comparison to existing methods.
This result demonstrates that 4D patch-level interaction features are capable of extracting the appropriate clues for accurate gaze target detection.
Besides, our method is a novel architecture based mainly on encoders (relative decoder parameters less than 1\%).
We conduct a comprehensive comparison of the parameter distribution in ViTGaze with those of existing SOTA methods as illustrated in Tab.~\ref{tab:decoders}.
Specifically, we list the parameter distributions of both the SOTA single-modality (RGB-only) method~\cite{Tu_HGGTR_2022_CVPR} and the SOTA multi-modality method~\cite{Tonini_GOT_2023_ICCV}.
The top-performing RGB-only method employs a hybrid architecture as the encoder for feature extraction, and it adopts a decoder constituting 26\% of the total parameter count for facilitating further information processing.
The most effective multi-modality method incorporates an additional object decoder constituting 18\% of the total parameter count for extra object detection.
This, in turn, leads to a two-stage decoder structure that accounts for 40\% of the total parameter count.
Our method harnesses the robust representational capabilities of pre-trained ViT, requiring only a prediction head that constitutes less than 1\% of the total parameter count to accomplish gaze following.
Additionally, pre-training allows us to achieve more robust interaction information extraction with a transformer-based encoder that occupies only 70\% of the parameter count compared with other methods.
This demonstrates that the complex and heavy design may not be necessary for gaze following. 

\begin{table*}[t]
    \centering
    \tablestyle{20pt}{1.1}
    \caption{Comparison of parameters. \thename{} has the lightest decoder, compared with the previous SOTA single-modality method~\protect\cite{Tu_HGGTR_2022_CVPR} and multi-modality method~\protect\cite{Tonini_GOT_2023_ICCV}.}
    \begin{tabular}{llccc}
    \toprule
    Methods & Components & \#Params. & AUC$\uparrow$ & Min. Dist.$\downarrow$ \\
    \midrule
    \multirow{2}{*}{Tu et al.~\cite{Tu_HGGTR_2022_CVPR}} & Encoder & 32 M & \multirow{2}{*}{0.917} & \multirow{2}{*}{0.069} \\ & Decoder & 11 M \\
    \midrule
    \multirow{3}{*}{Tonini et al.\cite{Tonini_GOT_2023_ICCV}} & Encoder & 32 M & \multirow{3}{*}{0.922} & \textbf{\multirow{3}{*}{0.029}} \\ & Object decoder & 10 M \\ & Gaze decoder & 12 M \\
    \midrule
    \multirow{2}{*}{ViTGaze (Ours)} & Encoder & \textbf{22 M} & \textbf{\multirow{2}{*}{0.949}} & \multirow{2}{*}{0.047} \\ & Decoder & \textbf{$\ll$1 M} \\
    \bottomrule
    \end{tabular}
    \label{tab:decoders}
\end{table*}

\subsection{Ablation Study}
In this section, we conduct experiments on the GazeFollow~\cite{Recasens_GazeFollow_2015_NIPS} benchmark to validate the effectiveness of our proposed method.
All the experiments are conducted with ViT-small~\cite{dinov2} as the backbone.

\begin{table}[t]
    \centering
    \tablestyle{8.5pt}{1.1}
    \caption{Ablation of multi-level 4D features. Multi-level and multi-head attention maps, which we refer to as multi-level 4D features, from pre-trained ViTs significantly improve gaze heatmap prediction. 2D represents image token features, S-4D represents 4D features from single level, and M-4D represents 4D features from multiple levels.}
    \begin{tabular}{lrrr}
        \toprule
        Features & AUC$\uparrow$ &  Min. dist.$\downarrow$ & Avg. dist.$\downarrow$ \\
        \midrule
        2D features     & 0.903    & 0.610   & 0.700    \\
        S-4D features   & 0.939    & 0.318   & 0.398    \\
        \rowcolor{gray!20}
        M-4D features   & \textbf{0.949}    & \textbf{0.047}   & \textbf{0.105}    \\
        \bottomrule
    \end{tabular}
    \label{tab:4dinteraction}
\end{table}

\begin{table}[t]
    \centering
    \tablestyle{5pt}{1.1}
    \caption{Ablation of spatial guidance and the auxiliary head. SG and AH stand for the 2D spatial guidance module and the auxiliary head, respectively.}
    \begin{tabular}{lrrr}
        \toprule
        Aggregation methods & AUC$\uparrow$ & Min. dist.$\downarrow$ & Avg. dist.$\downarrow$ \\
        \midrule
        Max. pooling   & 0.941             & 0.054            & 0.114             \\
        Avg. pooling   & 0.942             & 0.055            & 0.115             \\
        SG             & 0.938             & 0.086            & 0.154             \\
        \rowcolor{gray!20}
        SG+AH          & \textbf{0.949}    & \textbf{0.047}   & \textbf{0.105}    \\
        \bottomrule
    \end{tabular}
    \label{tab:pam}
\end{table}

\textbf{Multi-level 4D interaction features.}
To demonstrate the effectiveness of multi-level 4D interaction features (M-4D features) for gaze following, we conduct a study on the features used for gaze prediction.
We build a variant via 2D feature maps i.e. the final output of the ViT for prediction. 
We also compare our method to a single-level variant (S-4D features) that uses only attention maps in the last block to verify the effect of multi-level relation fusion.
The detailed results of the study on GazeFollow are presented in Tab.~\ref{tab:4dinteraction}.
According to the experimental results, 4D interaction features are more effective in capturing person-scene relations, whereas 2D features are incapable of capturing patch interactions and cannot directly determine the gaze targets in an end-to-end training paradigm.
Furthermore, our multi-level approach outperforms its single-level counterpart, highlighting the significance of capturing relationships at different levels.
Single-level 4D interactions fail to provide adequate local representation and cannot predict precise gaze targets.

\vspace{0.5em}
\textbf{2D spatial guidance.}
A comparative study is conducted to validate the significance of 2D spatial guidance, as illustrated in Tab.~\ref{tab:pam}.
Direct pooling of the interaction features in the head region achieves an acceptable result, indicating that it is unnecessary for the complex design of head feature extraction.
Moreover, the proposed 2D spatial guidance achieves a 0.7\% improvement in the AUC, demonstrating that our design results in more informative fusion to further increase the performance.
Furthermore, the introduction of the auxiliary head prediction task effectively supervises this module and increases the effectiveness of training the 2D spatial guidance, consequently enhancing the performance of spatial guidance.

\textbf{Pre-training.}
To validate the significance of pre-trained ViT for interaction feature extraction, we conduct experiments using DeiT~\cite{deit}, DINO~\cite{dino}, iBOT~\cite{zhou2021ibot}, and DINOv2~\cite{dinov2} as the backbone on GazeFollow~\cite{Recasens_GazeFollow_2015_NIPS}, as illustrated in Tab.~\ref{tab:pretraining}.
DINO, iBOT, and DINOv2 are pre-trained on a large amount of unlabelled data to learn robust visual representations with different self-supervising algorithms.
DeiT is a supervised learning method that serves as a benchmark for comparison and contrast of the outcomes and effectiveness of self-supervised methods.
DeiT, DINO, and iBOT are trained on the ImageNet dataset~\cite{imagenet}, whereas DINOv2 is trained on the larger LVD-142M dataset~\cite{dinov2}.
Compared with DeiT pre-training, ViT with self-supervised pre-training can greatly improve the performance of the model.
The gaze following performance of these methods is highly consistent with their results on ImageNet for representation learning.
Compared with the pre-trained ViT with DeiT pre-training, the model utilizing DINOv2 pre-trained weights achieves a notable 6.8\% improvement in AUC.
In Appendix~\ref{sec:backbone}, the visualization results of ViTGaze built upon different pre-training methods are presented to illustrate the impact of pre-training methods on the granularity of predictions.
The experimental results demonstrate that the robust representational capabilities of pre-trained ViT effectively enhance gaze following performance.

\textbf{Loss weights.} The loss weights, i.e. $\lambda_1=100$, and $\lambda_2=1$, are based on the methodology outlined in Ref.~\cite{Chong_VideoAttn_2020_CVPR}.
We conduct experiments on $\lambda_2$ and $\lambda_3$ in Tab.~\ref{tab:lambda1}.
The results demonstrate that the ratio of $100: 1: 1$ keeps each loss at the same scale to ensure that no single task dominates the training process and therefore achieves the best performance.

\begin{table}[t]
    \centering
    \tablestyle{9.5pt}{1.1}
    \caption{Ablation of pre-training. Self-supervised pre-training effectively enhances the ability of the model to extract interaction information.}
    \begin{tabular}{lrrr}
        \toprule
        Pre-training & AUC$\uparrow$ & Min. dist.$\downarrow$ & Avg. dist.$\downarrow$ \\
        \midrule
        DeiT\cite{deit}   & 0.881    & 0.189   & 0.267    \\
        DINO\cite{dino}   & 0.923    & 0.086   & 0.154    \\
        iBOT\cite{zhou2021ibot}   & 0.934    & 0.069   & 0.133    \\
        \rowcolor{gray!20}
        DINOv2\cite{dinov2} & \textbf{0.949}    & \textbf{0.047}   & \textbf{0.105}    \\
        \bottomrule
    \end{tabular}
    \label{tab:pretraining}
\end{table}

\begin{table}[t]
    \centering
    \tablestyle{6.5pt}{1.1}
    \caption{Ablation of loss weights. The adopted loss settings strike the balance of each task and achieve the highest performance. $\lambda1$, $\lambda2$, and $\lambda3$ are weights of gaze heatmap loss, gaze in-out loss, and auxiliary head regression loss.}
    \begin{tabular}{lllrrrr}
        \toprule
        $\lambda_1$  & $\lambda_2$ & $\lambda_3$ & AUC$\uparrow$ & Min. dist.$\downarrow$ & Avg. dist.$\downarrow$ \\
        \midrule
        100 & 1.0   & 0.1    & 0.945    & 0.050   & 0.109    \\
        100 & 1.0   & 10.0   & 0.943    & 0.057   & 0.120    \\
        100 & 0.1   & 1.0    & 0.948    & 0.049   & 0.108    \\
        100 & 10.0  & 1.0    & 0.941    & 0.058   & 0.121    \\
        \rowcolor{gray!20}
        100 & 1.0   & 1.0    & \textbf{0.949}   & \textbf{0.047}   & \textbf{0.105}    \\
        \bottomrule
    \end{tabular}
    \label{tab:lambda1}
\end{table}

\begin{figure*}[ht]
    \centering
    \includegraphics[width=17cm]{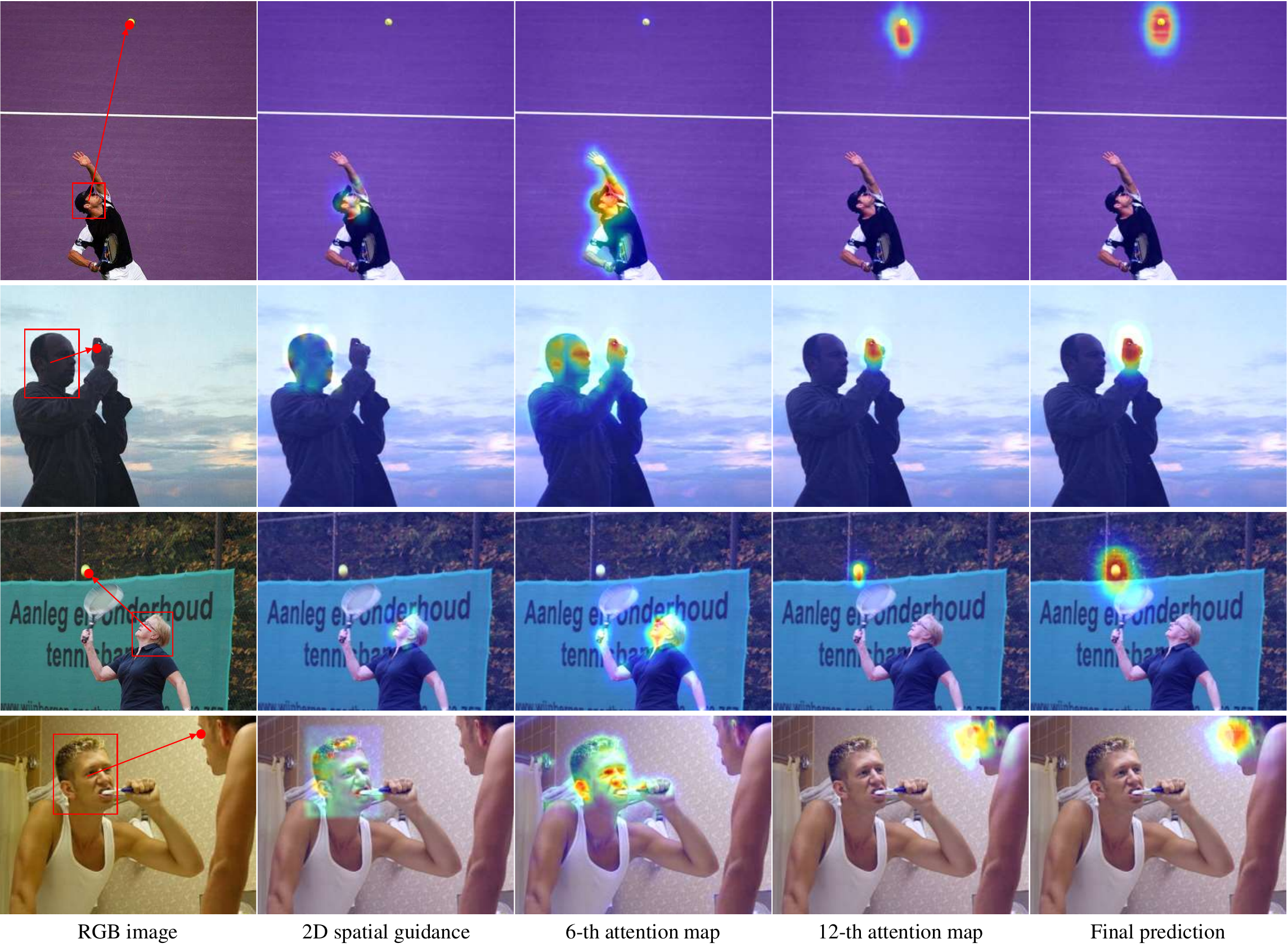}
    \caption{Visualization of main components of \thename{} on the GazeFollow~\cite{Recasens_GazeFollow_2015_NIPS} dataset. We observe that the 2D spatial guidance (the 2-nd column) successfully extracts the patches overlapped with the head, and the multi-level interaction maps (the 3-rd and 4-th columns) reflect the region interactions. Specifically, the 6-th layer captures local regions, whereas the 12-th layer has an advanced capacity for global modeling.}
    \label{fig:visualization}
\end{figure*}

\subsection{Visualization Results}
The visualization results on the GazeFollow dataset~\cite{Recasens_GazeFollow_2015_NIPS} are provided in Fig.~\ref{fig:visualization}, where the 2D spatial guidance, multi-level person-specific interaction features, and the final predictions are displayed.
We observe that the 2D spatial guidance visualized in the second column extracts the patches overlapped with the key points of the face of a person.
The interaction features generated by the medium layer (the 6-th transformer block) reflect the local interactions of the person's head and body and highlight the key points such as eyes.
The interaction features of the final layer in the 3-rd column show the global interactions and roughly indicate the gaze targets. 
These observations reflect the importance of multi-level interaction features and 2D spatial guidance as described above.
The visualization results on the VideoAttentionTarget dataset~\cite{Chong_VideoAttn_2020_CVPR} are provided in Appendix~\ref{sec:vis}.

\begin{figure*}[t]
    \centering
    \includegraphics[width=17cm]{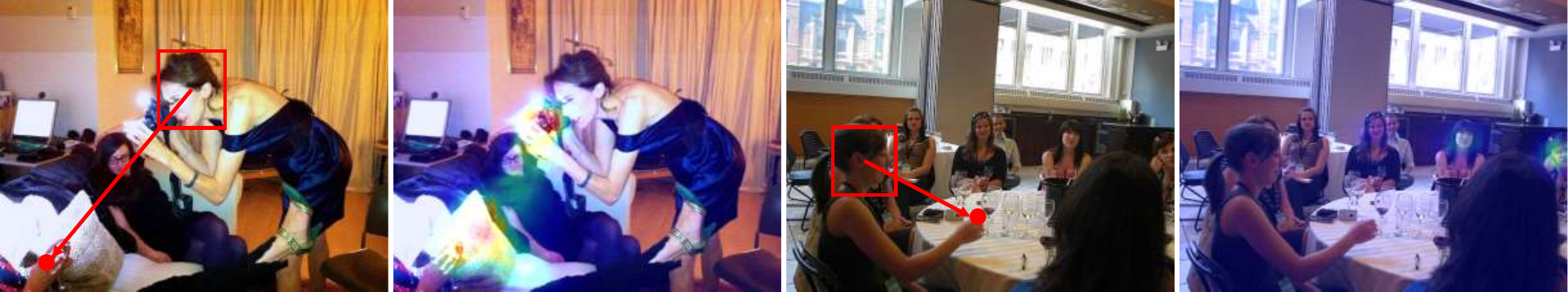}
    \caption{Failure cases of \thename{}. The predictions may become ambiguous in occluded situations.}
    \label{fig:failure}
\end{figure*}

\section{Conclusion}\label{sec5}
In this paper, we propose a new high-performance single-modality gaze following framework, \thename{}, which is based on pre-trained plain vision transformers.
It consists of a 4D interaction encoder, 2D spatial guidance information, and two lightweight predicting heads.
Our key observation is that inter-token interactions within self-attention can be transferred to interactions between humans and scenes.
\thename{} presents a brand-new paradigm of gaze following mainly based on encoders and its relative decoder parameters is less than 1\%.
This first demonstrates that a concise single-modality framework could also achieve high gaze following performance with the help of pre-trained vision transformers. 
We hope that our work can inspire more research on gaze following with pre-trained vision transformers.

The limitations and future work are as follows.
As illustrated in Fig.~\ref{fig:failure}, in practical situations, the predictions of \thename{} may become ambiguous, particularly when individuals engage in complex environments with occlusions.
To address this issue, it would be beneficial to integrate the prior of gaze dynamics and global saliency.
We leave them as our future work.

\begin{appendices}

\section{Detailed training configurations}\label{sec:settings}
We summarize the detailed training configurations on GazeFollow~\cite{Recasens_GazeFollow_2015_NIPS} in Tab.~\ref{tab:gf_settings}.
We follow DINOv2~\cite{dinov2} to increase the resolution of images to $518\times 518$ in the last epoch.
During our training process, we apply several data augmentation techniques to enhance the dataset.
These techniques include head bounding box jittering, color jittering, random resizing and cropping, random horizontal flips, random rotations, and random masking of the scene patches.

For the VideoAttentionTarget dataset~\cite{Chong_VideoAttn_2020_CVPR}, we perform fine-tuning on the model pre-trained on the GazeFollow dataset.
During this fine-tuning phase, we do not adopt a learning rate scheduler and opt for a fixed learning rate of $1.0\times 10^{-4}$, which is sustained for a single training epoch.
All other training configurations remained unchanged from those applied during training on the GazeFollow dataset.

\section{Comparison of pre-training methods}\label{sec:backbone}
\begin{figure*}[t]
    \centering
    \includegraphics[width=\textwidth]{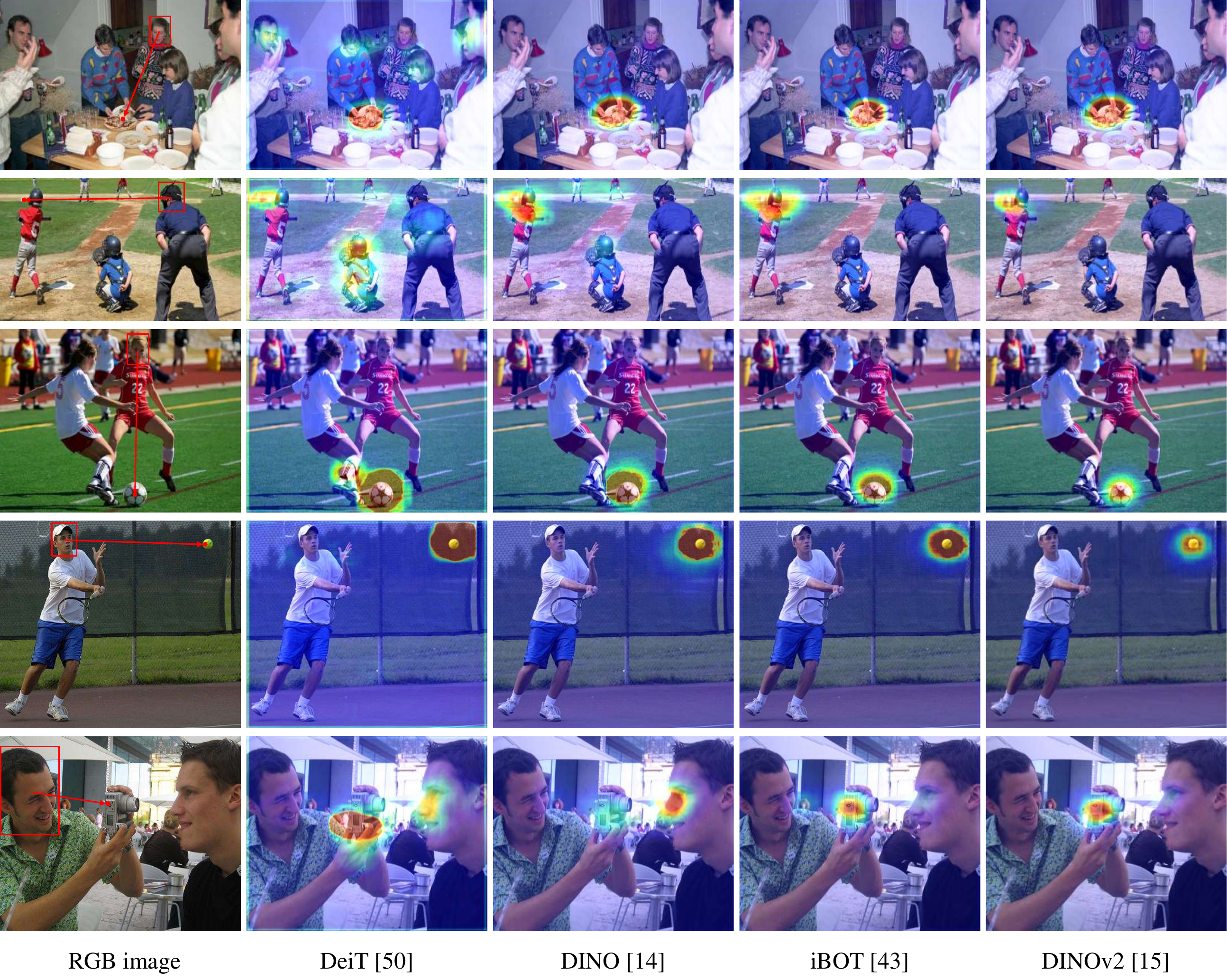}
    \caption{Comparison of pre-training methods. The models built upon self-supervised pre-trained ViTs perform more accurately and finer-grained gaze target predictions.}
    \label{fig:backbones}
\end{figure*}

To highlight the significance of pre-trained ViTs, we visualize the results of ViTGaze built upon different pre-training methods on the GazeFollow dataset~\cite{Recasens_GazeFollow_2015_NIPS}, as shown in Fig.~\ref{fig:backbones}.
The supervised pre-trained method~\cite{deit}, as observed in our evaluation, tends to yield a greater number of false positive masks within the salient regions that lie outside of the individual's field of view.
In contrast to supervised pre-training, the utilization of self-supervised pre-trained ViTs~\cite{dino,zhou2021ibot,dinov2} substantially enhances the gaze following performance.
Furthermore, the performance of the models consistently aligns with the outcomes of their representation learning experiments on the ImageNet benchmark.
Among all the pre-training techniques at our disposal,
the model built upon the pre-trained ViT through the DINOv2 pre-training method stands out because it achieves the highest levels of precision and granularity in predicting gaze targets.

\begin{table}[t]
\centering
\tablestyle{15pt}{1.1}
\caption{Training configurations on the GazeFollow dataset.}
\begin{tabular}{lc}
\toprule
Config & Value \\
\midrule
optimizer & AdamW~\cite{adamw} \\
optimizer momentum decay rate & 0.9 \\
optimizer variance decay rate & 0.99 \\
optimizer epsilon & $1.0\times10^{-8}$ \\
weight decay & 0.1 \\
learning rate & 0.01 \\
learning rate schedule & cosine decay \\
layer-wise learning rate decay & 0.65 \\
warmup iteration ratio & 0.01 \\
training epochs & 15 \\
batch size in total & 48 \\
\bottomrule
\end{tabular}
\label{tab:gf_settings}
\end{table}

\section{Visualization Results on the VideoAttentionTarget Dataset}\label{sec:vis}
\begin{figure*}[t]
    \centering
    \includegraphics[width=\textwidth]{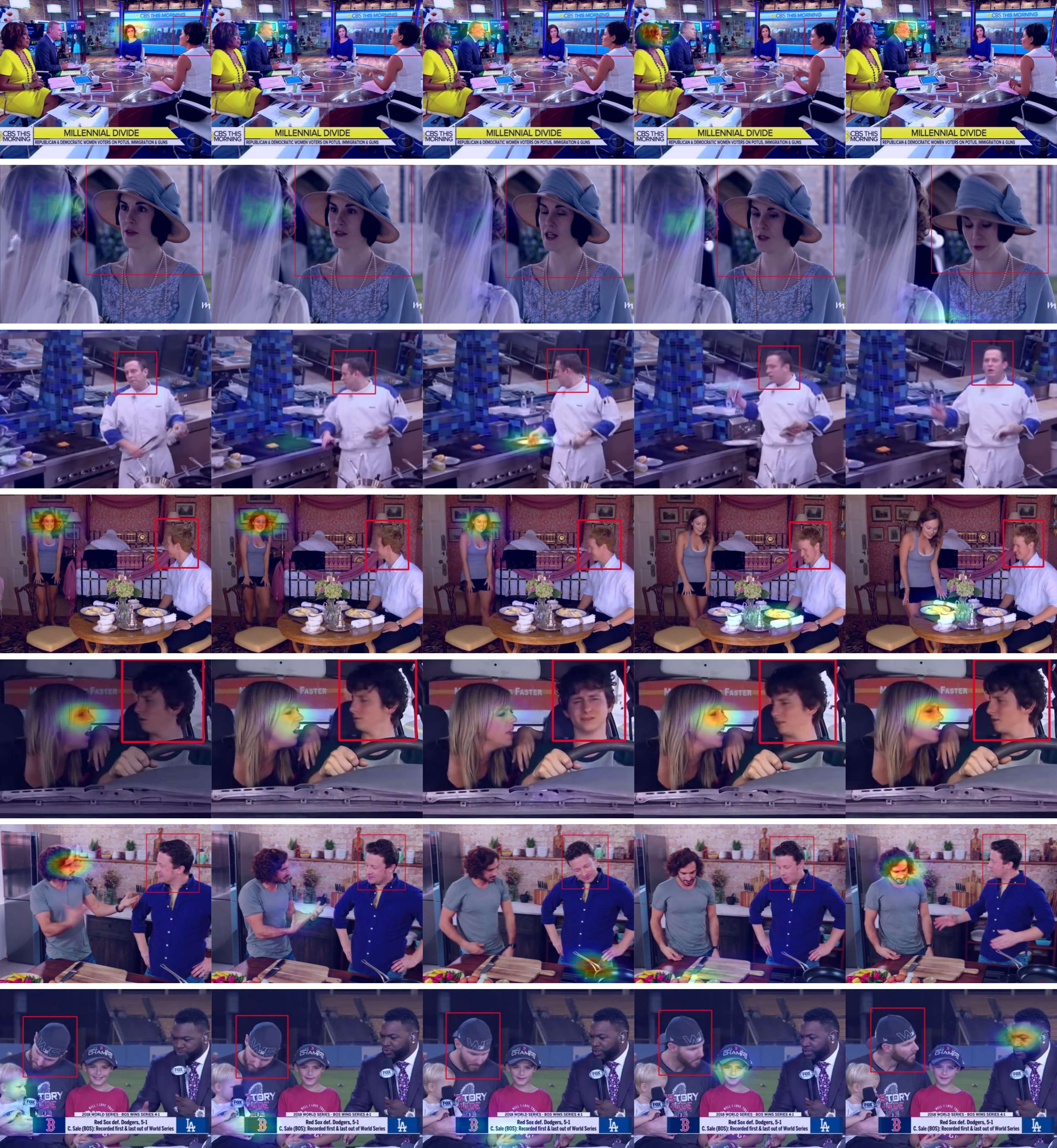}
    \caption{Visualization results on the VideoAttentionTarget~\cite{Chong_VideoAttn_2020_CVPR} dataset.}
    \label{fig:vat_visualize}
\end{figure*}

We visualize the results on the VideoAttentionTarget dataset~\cite{Chong_VideoAttn_2020_CVPR}, as illustrated in Fig.~\ref{fig:vat_visualize}.
For each video clip, we extract one image at intervals of eight frames and subsequently select a consecutive sequence of five extracted images for presentation and analysis.

Notably, even in the absence of explicit time-series modeling, our approach remarkably excels at effective gaze following in video sequences.
This noteworthy observation demonstrates that ViTGaze constructed upon pre-trained ViT can extract abundant interaction features from RGB images without any additional input modality, facilitating robust gaze following.
Another intriguing observation is that, despite the decoupling of gaze heatmap prediction and watching-outside prediction in our method,
the responses generated on the predicted heatmap effectively reflect the confidence of the model in the direction where the person is looking.
The predicted heatmap no longer exhibits discernible activations when the gaze target is outside of the frame.

\end{appendices}

\paragraph{Abbreviations}
AP, average precision; AUC, area under the curve; GT, ground truth; MLP, multi-layer perception; SOTA, state of the art; ViT, vision transformer.

\section*{Declarations}

\paragraph{Availability of data and material}
Our code is available at \url{https://github.com/hustvl/ViTGaze}. The \\ datasets analyzed during the current study are available in GazeFollow~\cite{Recasens_GazeFollow_2015_NIPS} and VideoAttentionTarget~\cite{Chong_VideoAttn_2020_CVPR}.

\paragraph{Competing interests}
The authors declare no competing interests.

\paragraph{Funding}
This work was supported by the National Science and Technology Major Project (No. 2022YFB4500602).

\paragraph{Author contribution}
YS conceived the initial ideas, conducted detailed experiments, and drafted the paper.
XW revised the manuscript and improved the research ideas.
JY revised the manuscript and improved the experimental design.
WL, JZ, and XX systematically refined the research framework and guided the writing of the paper. 
All the authors read and approved the final manuscript.

\bibliographystyle{unsrt}
\bibliography{reference}

\end{document}